\documentclass[a4paper,twoside]{article}

\usepackage{epsfig}
\usepackage{subcaption}
\usepackage{calc}
\usepackage{amssymb}
\usepackage{amstext}
\usepackage{amsmath}
\usepackage{amsthm}
\usepackage{multicol}
\usepackage{pslatex}
\usepackage{apalike}
\usepackage{stmaryrd}
\graphicspath{{images/}}
\usepackage[ruled,vlined]{algorithm2e}
\DontPrintSemicolon
\usepackage{multirow}
\usepackage[switch]{lineno} 
\usepackage{color}
\usepackage{dblfloatfix}
\usepackage{SCITEPRESS}     

\usepackage{hyperref}

\newenvironment{myalign}%
  {\linenomath\align}%
  {\endalign\endlinenomath}
\begin{document}


\title{Dense open-set recognition with \\  synthetic outliers generated by Real NVP}

\author{\authorname{Matej Grcić, Petra Bevandić and Siniša Segvić}
\affiliation{Faculty of Electrical Engineering and Computing, University of Zagreb, Croatia}
\email{\{matej.grcic, petra.bevandic, sinisa.segvic\}@fer.hr}
}



\abstract{Today's deep models are often unable to detect inputs
which do not belong to the training distribution.
This gives rise to confident incorrect predictions 
which could lead to devastating consequences
in many important application fields such as healthcare and autonomous driving.
Interestingly, both discriminative and generative models
appear to be equally affected.
Consequently, this vulnerability
represents an important research challenge.
We consider an outlier detection approach 
based on discriminative training 
with jointly learned synthetic outliers.  
We obtain the synthetic outliers 
by sampling an RNVP model 
which is jointly
trained to generate datapoints 
at the border of the training distribution.
We show that this approach 
can be adapted for simultaneous 
semantic segmentation and dense outlier detection.
We present image classification experiments on CIFAR-10,
as well as semantic segmentation experiments 
on three existing datasets 
(StreetHazards, WD-Pascal, Fishyscapes Lost \& Found),
and one contributed dataset. 
Our models perform competitively 
with respect to the state of the art 
despite producing predictions with only one forward pass.
}

\onecolumn \maketitle \normalsize \setcounter{footnote}{0} \vfill

\section{\uppercase{Introduction}}

Early computer vision workflows involved 
handcrafted feature engineering 
and shallow discriminative models.
However, despite hard work 
and notable scientific contribution,
the resulting features 
\cite{Lowe04,SanchezPMV13,RostenD06}
were insufficient in many computer vision tasks. 
Emergence of deep convolutional models \cite{KrizhevskySH12} 
enabled implicit feature engineering. 
Consequently, computer vision research is now focused 
on designing deep models which are trained in end-to-end fashion
\cite{SimonyanZ14a,HeZRS16,HuangLMW17}.

Today, deep models deliver state-of-the-art performance 
in most computer vision tasks.
However, whenever there is a domain shift 
between the train and the test distributions,
we witness overconfidence 
of incorrect predictions
\cite{Lakshminarayanan17,GuoPSW17}. 
This issue poses a direct threat 
to the safety of AI-based systems
for healthcare \cite{Yingda2020}, 
autonomous driving \cite{ZendelHMSD18} 
and other critical application fields.
Therefore, open-set recognition \cite{BendaleB15} becomes 
an increasingly important research objective.
The desired models should be able to detect
foreign samples while also fulfilling 
their primary discriminative task.
This problem has been also addressed in several related settings such as
anomaly detection \cite{Andrews2016TransferRF},
uncertainty estimation \cite{MalininG18}, and out-of-distribution (OOD) detection \cite{HendrycksG17}.
We focus on open-set recognition
and note that the proposed approach can be generalized to other settings.

Formally, OOD detection can be viewed as a form of
binary classification where the model has to predict 
whether a given sample belongs to the training dataset.
In practice, this can be achieved by designing the model to predict an OOD 
score given the input.
We assume that the score function $s(\textbf{x}): \textbf{X} \rightarrow \mathbb{R}$
assigns OOD score to any given sample. 
A convenient baseline approach \cite{HendrycksG17}
expresses the score function
in terms of maximum softmax probability: 
$s_{MSP}(\textbf{x}) = 1 - \max_c P_\theta(c|\textbf{x})$.
The score function can also be expressed with softmax entropy:
$s_H(\textbf{x})=-\sum_i P_\theta(c_i|\textbf{x}) \, \mathrm{log} \, P_\theta(c_i|\textbf{x})$ \cite{HendrycksMD19}.
We evaluate our dense OOD detection approach both with $s_{MSP}(\textbf{x})$
and $s_H(\textbf{x})$.

Typically, we wish to achieve open-set recognition with a single
forward pass in order to promote cross-task synergy and
allow real-time inference.
Unfortunately, the two tasks may negatively affect each other
since this a form of the multi-objective optimization.

This paper presents an open-set recognition approach 
based on jointly learned synthetic examples generated at the border
of the training distribution \cite{LeeLLS18}.
Instead of adversarial generation \cite{goodfellow14nips}, 
we prefer to base our approach on 
invertible normalized flow \cite{DinhSB17}
due to better distribution coverage \cite{LucasSASV19} 
and opportunity to generate images of arbitrary resolution.
The resulting method outperforms GANs 
in equivalent image-wide experiments.
We argue that the proposed approach 
is especially suitable for adaptation to dense prediction
due to capability of normalizing flows to generate images of
arbitrary size.
We evaluate our models on several
datasets for dense open-set recognition
and demonstrate competitive performance 
with respect to the state of the art.

\section{\uppercase{Related work}}

Previous work
covers various facets of OOD detection.
As usual in computer vision research,
datasets and benchmarks are 
critical research tools
since they help us identify effective approaches.
Generative models are important for our work
since they allow us to generate 
synthetic training samples
for discriminative OOD detection.

\subsection{Out-of-distribution Detection}
The most popular OOD detection approaches include
likelihood evaluation \cite{NalisnickMTGL19},
assesing the prediction confidence \cite{devriesConf},
exploiting an additional negative dataset \cite{HendrycksMD19},
and
modification of the loss function \cite{LeeLLS18}.
In theory, OOD detection can be elegantly implemented
by evaluating the likelihood of a given sample with a
suitable generative model.
A generative model capable of exact likelihood evaluation should assign OOD samples a lower likelihood.
However, \cite{NalisnickMTGL19} and \cite{SerraAGSNL20} show that generative models
with exact likelihood evaluation tend to assign simple outliers a higher likelihood than to complex inliers.
Exhaustive analysis shows that flow-based generative models
are not suitable for this task due to their architecture  \cite{Kirichenko2020}.
Still, \cite{GrathwohlWJD0S20} manage to detect outliers using the
gradient of sample likelihood, while \cite{RenLFSPDDL19}
detect outliers by evaluating likelihood ratios of two generative models.
We do not use these approaches since our primary task is open-set semantic segmentation
which requires specifically designed discriminative models
for dense prediction.

OOD detection can also be expressed throughout
a modification of the model architecture.
This has been attempted by adding 
a confidence prediction head 
which is supposed to emit 
low confidence prediction
for OOD samples \cite{devriesConf}.
However, learned confidence is able to account 
only for difficulties which have been seen in the training data (aleatoric uncertainty), 
while remaining mostly unable to manage OOD inputs (epistemic uncertainty).

A preprocessing approach known as ODIN \cite{LiangLS18} 
does not require any change in model's training procedure.
They add a small perturbation to the input 
which leads to better separation 
between in- and out-of-distribution samples.
However, this approach results in only slight improvements 
over the max-softmax baseline \cite{bevandic19simul}.
Additionally, it requires multiple passes trough the model,
which considerably increases the inference latency.

Discriminative OOD detection can be improved by
utilizing a diverse negative dataset \cite{HendrycksMD19,bevandic19simul}.
Additionally, outliers can be detected by observing
the distributional uncertainty 
of prior networks \cite{MalininG18}.
We differ from these approaches, 
since we do not require 
training on negative data.

The negative dataset can be replaced 
by an adversarial network 
which generates artificial outliers \cite{LeeLLS18,nitsch2020out}.
This setup requires that the discriminative classifier output
uniform distribution in synthetic samples.
Consequently, the generator network 
receives a signal from classifier's loss
which moves generated samples 
to the border of the training distribution.
Our approach is closely related with this method
and hence we present an empirical comparison in the experiments.
The main differences are that
i) our method uses RNVP instead of GAN which allows us to generate outliers of different sizes, 
and ii) we present an adaptation for dense open-set recognition.

\subsection{Dense Open-set Recognition}

OOD detection can be combined with pixel-level classification to
achieve dense open-set recognition.
Testing open-set performance 
of dense discriminative models 
proved to be a difficult task. 
Available benchmarks fail to fully cover 
all open-world situations
but still pose a challenge to present models. 
However, evaluating the model on multiple benchmarks 
can give us a better notion on open-set recognition performance. 
The WildDash benchmark \cite{ZendelHMSD18} 
evaluates capability of the model 
to deal with challenging real-world situations 
and outright negative images.
We do not evaluate directly on this dataset since it does not include test images with  mixed content (both inliers and outliers).
The Fishyscapes Lost \& Found benchmark \cite{fs-lf}
evaluates the ability to detect small OOD objects
on the road surface.
The StreetHazards dataset \cite{bddanomaly} 
includes synthetic images created by the Unreal Engine, 
while the BDD-Anomaly dataset \cite{bddanomaly} is created from the 
Berkley Deep Drive dataset (BDD) \cite{bdddataset} 
by selecting \textit{motorcycle} and 
\textit{train} classes as anomalies.
The WD-Pascal dataset \cite{bevandic19simul}
contains WildDash images with pasted animals 
from the Pascal VOC \cite{EveringhamGWWZ10} dataset.

There are several prior approaches 
which adress dense open-set recognition.
Epistemic uncertainty attempts to discern uncertainty
due to insufficient knowledge from uncertainty
due to insufficient supervision \cite{KendallG17}.
However, their approach assumes that MC dropout corresponds
to Bayesian model sampling which may not be satisfied in practice.
Additionally, Bayesian model sampling is unable 
to account for distributional shift \cite{MalininG18}.
OOD detection can also be framed as comparison 
between the original image and its sythesized version 
which is conditionally generated 
from the predictions \cite{Yingda2020}.
However image-to-image translation is a hard problem;
OOD input is not a necessary condition for getting a poor reconstruction.
Employing confidence of multiple DNNs trained in one-vs-all setting \cite{Franchi2020}
currently achieves the best OOD detection results on StreetHazards dataset.
All these approaches require multiple forward passes through complex models and are therefore much slower than our approach.

Recognition of OOD input can also be improved by training
on several positive and negative datasets.
However that requires extraordinary efforts
for making different datasets mutually compatible \cite{MSeg_2020_CVPR}.
A more affordable approach assumes one specialized inlier 
dataset and one general purpose noisy negative dataset \cite{bevandic19simul}.
Different than both these approaches,
we do not use negative training data.
Instead we jointly train
a generative model for synthetic negative samples 
which we use for training of the discriminative model.
Consequently, our approach does not incur any bias 
due to particular choice of the negative dataset.

\subsection{Generative Modeling}

Many generative models approximate the data distribution $p_D$
using the model distribution $p_\theta$ defined by:
\begin{equation} 
  \label{gen-mod}
    p_{\theta}(\textbf{x}) = \frac{p'_\theta(\textbf{x})}{Z}, \quad Z=\int_\textbf{x} p'_\theta(\textbf{x}) dx.
\end{equation}
In the previous equation, $p'_\theta$ denotes the unnormalized distribution
modeled with a deep model parameterized with $\theta$,
while $Z$ is a normalization constant.

Restricted Boltzmann machines (RBM) \cite{rbm-paper} learn
the data distribution $p_D$ by utilizing a two-phase training procedure.
The positive phase increases the value of $p'_\theta(\textbf{x})$ for every datapoint $\textbf{x}_i$,
while the negative phase updates the normalization constant $Z$ in order to keep
$p_\theta(\textbf{x})$ a valid distribution.

Invertible normalizing flows \cite{DinhKB14} are trained by likelihood maximization which essentially
increases the value of $p_\theta(\textbf{x})$ for every datapoint $\textbf{x}_i$.
This corresponds to the positive phase of RBM's training.
Different than RBM's, normalizing flows require that
the nonlinear transformation is bijective.
In particular, normalizing flows consider 
an invertible transformation
$f_\theta: \textbf{X} \rightarrow \textbf{Z}$ 
which maps the desired complex data distribution 
to a simpler latent distribution.
Consequently, the change of variable formula:
$p_\theta(\textbf{x}) = p_z(f_\theta(\textbf{x})) |\mathrm{det}\frac{\partial f_\theta(\textbf{x})}{\partial \textbf{x}}|$
can be applied.
Since we transform normalized latent distribution 
with $f_\theta$, which is bijective by design,
the change of variables gives us a guarantee
that the $p_\theta(\textbf{x})$ will remain normalized.
Hence, the negative phase for the flow-based models is unnecessary.
Real NVP (RNVP) \cite{DinhSB17} is a flow-based generative model
which captures the dataset distribution 
by learning to bijectively transform it with
a powerful set of affine transformations
to a simpler latent distribution such as a unit Gaussian.

\section{\uppercase{The Proposed Method}}

We approach the task of open set recognition 
by exploiting synthetic negative samples produced by
a RNVP model which is jointly trained with the open-set classifier.
We aim to improve OOD detection 
without harming recognition accuracy. 
We first apply the proposed setup in the image-wide setup
and later adapt for dense OOD detection. 

\subsection{Joint Training of Discriminative Classifier
  and Real NVP}
\label{sec:JTR}  
We assume that OOD performance of a classifier
can be improved by introducing an additional loss term 
that encourages it to emit the
uniform distribution in synthetic outliers
\cite{LeeLLS18}. 
This term can be expressed 
as Kullback-Leibler (KL) divergence 
between the model prediction in OOD samples
and the uniform distribution. 
The resulting compound classifier loss is:
\begin{myalign} \label{comp-loss}
   L_\mathrm{cls}(\theta_C) = 
   - &\mathbb{E}_{\hat{x}, \hat{y} \sim P_{in}} 
     [\mathrm{log} P_{\theta_C}(y = \hat{y} | \hat{x})] 
     \nonumber\\ 
     &+ 
   \lambda \cdot
   \mathbb{E}_{x \sim P_{out}} [ \mathrm{KL}(U || P_{\theta_C}(y|x))]
   \;.
\end{myalign}
Note that the distribution $P_{out}$ 
corresponds to synthetic outliers generated by RNVP.
Hence, the classifier loss encourages RNVP 
to generate data which is classified with high entropy. 
However, we also apply the standard negative log-likelihood loss to the RNVP model:
\begin{myalign}\label{rnvp-loss}
L_\mathrm{RNVP}(\theta_R) = 
  - \mathbb{E}_{\textbf{x} \sim P_{in}}  
    &[ \mathrm{log} \,  
      p_{\textbf{z}}(\textbf{f}_{\theta_R}(\textbf{x}))) 
      \nonumber\\  
      &+ 
      \mathrm{log} \, \left|\mathrm{det}\left(
        \frac{\partial \textbf{f}_{\theta_R}(\textbf{x})}
          {\partial \textbf{x}}\right)\right|]
  \; .
\end{myalign}
This loss encourages RNVP to generate the data 
which resembles the training dataset. 
The two losses ($L_\mathrm{cls}$ and $L_\mathrm{RNVP}$) are opposed,
but together they cause the RNVP 
to generate data 
which resemble inliers but get classified to uniform distribution.
Consequently, we colloquially state that our RNVP model 
generates samples at the border of the training distribution
\cite{LeeLLS18}. 

\begin{figure}[htb]
\centering
\includegraphics[width=0.95\linewidth]{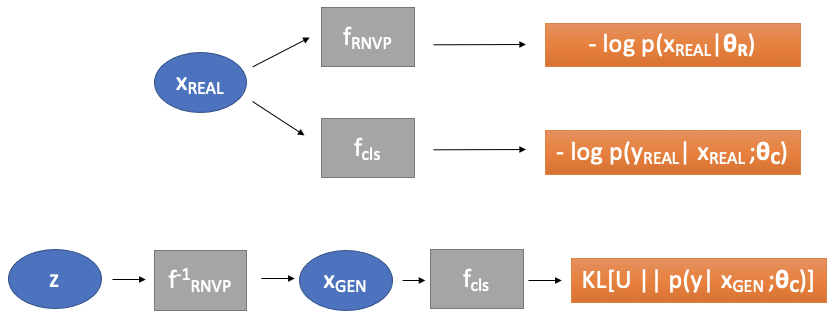}
\caption{
    Schematic overview of the proposed training procedure using losses (\ref{comp-loss}) and (\ref{rnvp-loss}).
  }
\label{fig:schema}
\end{figure}

Figure \ref{fig:schema} illustrates the proposed training procedure.
Real images are processed by the generative model $f_{RNVP}$
and by the discriminative model  $f_{cls}$.
The generative model outputs the likelihood 
which is subjected to generative NLL loss.
The discriminative model outputs posterior probability
which is subjected to cross-entropy with respect to the labels.
At the same time, a random latent vector $\mathbf{z}$ is drawn from a Gaussian distribution.
A synthetic outlier is obtained by sampling RNVP ($f^{-1}_{RNVP}$),
and processed by $f_{cls}$.
The resulting posterior is subjected to KL loss with respect to the uniform distribution. 
We summarize the joint training procedure in Algorithm \ref{alg-rnvp}.

\begin{algorithm}[h]
    \textbf{Require:} $\lambda > 0$\;
    \textbf{Define} RNVP: $\textbf{z} = \textbf{f}_{\theta_R}(\textbf{x}),  \textbf{x} = \textbf{f}_{\theta_R}^{-1}(\textbf{z})$\;
    \textbf{Define} Classifier: $P_{\theta_C}(y|\textbf{x})$\;
    \textbf{Define} Optimizers: $O_\mathrm{R}(\theta_R), O_\mathrm{C}(\theta_R)$\;
    \Repeat{convergence}{
      \textbf{x}, y = obtain\_minibatch()\;
      \textbf{z} = sample $N(0, 1)$\;
      $L_\mathrm{cls} = -\mathrm{log} P_{\theta_C}(y|\textbf{x}) + \lambda \, \mathrm{KL}(U || P_{\theta_C}(y|\textbf{f}_{\theta_R}^{-1}(\textbf{z})))$\;
      $\theta_\mathrm{C} += O_\mathrm{C}.update(\nabla_{\theta_C} \, L_\mathrm{cls})$\;
      $L_\mathrm{RNVP} = -\mathrm{log}(p_{z}(\textbf{f}_{\theta_R}(\textbf{x}))) - \mathrm{log}\left(\left|det(\frac{\delta \textbf{f}_{\theta_R}(\textbf{x})}{\delta \textbf{x}})\right|\right)$\;
      $\theta_\mathrm{R} += O_\mathrm{R}.update(\nabla_{\theta_R} \, L_\mathrm{RNVP} + \nabla_{\theta_C} \, L_\mathrm{cls})$\;
    }
 \caption{Joint training of an open-set classifier and an RNVP model for generation of synthetic outliers.}
 \label{alg-rnvp}
\end{algorithm}

\subsection{Dense Open-set Recognition}
\label{sec:dense}

This section extends the described 
joint training of synthetic outliers
towards dense open-set recognition. 
Different than in the image-wide case 
where an image is either an inlier or an outlier,
dense OOD detection has to deal with images of mixed content. 
A real-world image may contain 
OOD objects
or exclusively consist of inlier content.
Thereby, outliers typically occlude the background,
which results in well-defined borders between the outlier and the inlier content.
We embed these observations into our dense open-set recognition approach as follows.
We jointly train RNVP
with a semantic segmentation model
as we described in \ref{sec:JTR}. 
Different than in the image-wide setup,
we paste the synthetic outliers 
provided by RNVP
at random locations within our regular training images. 
We train the dense prediction model 
to predict correct semantic segmentation in inlier pixels,
and the uniform distribution within the patches generated by RNVP.
The discriminative loss is defined by:
\begin{figure*}[h]
\begin{center}
\includegraphics[width=0.8\linewidth]{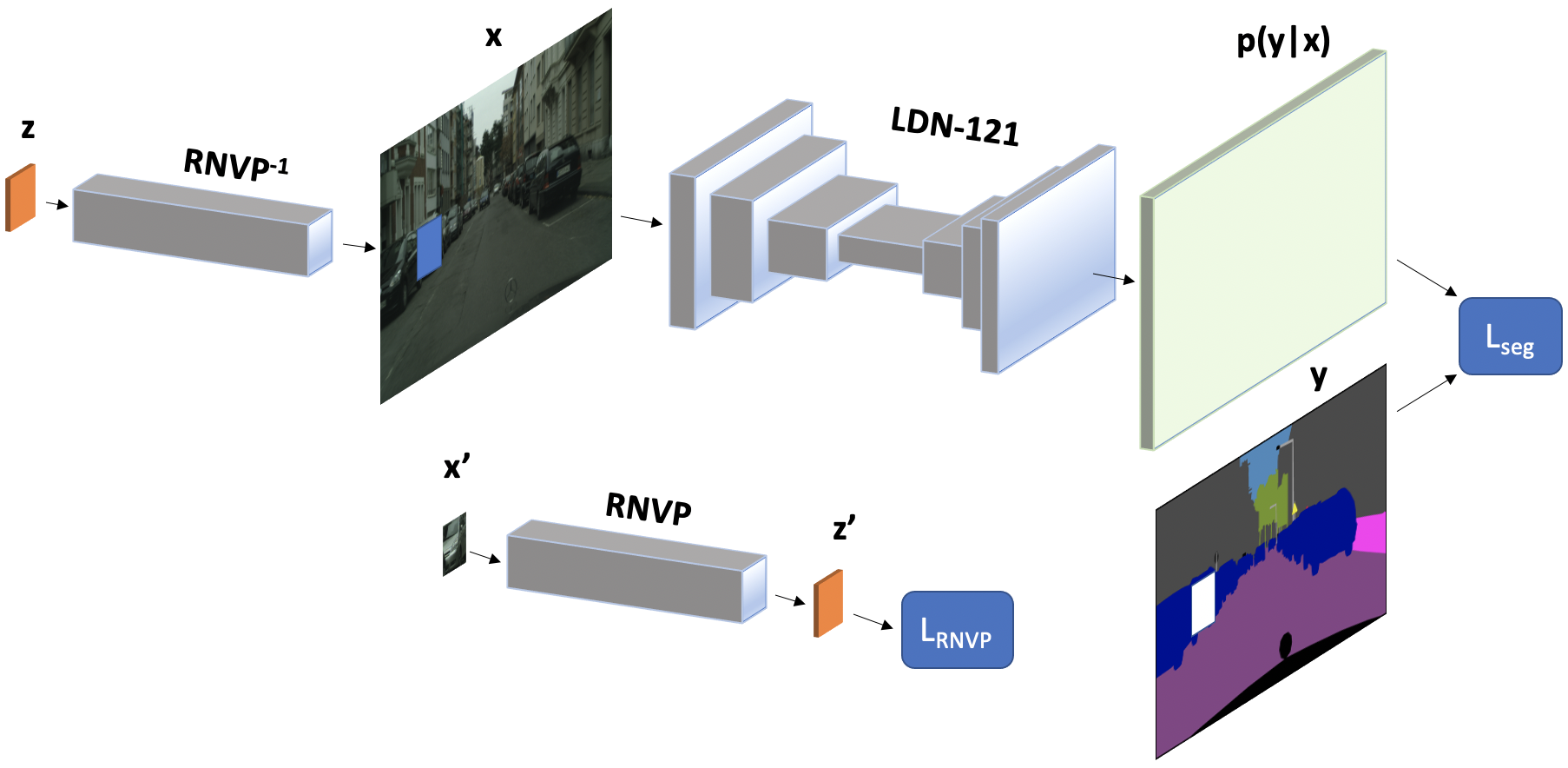}
\end{center}
   \caption{Schematic overview of the proposed training procedure for
   dense open-set recognition.
   The loss $L_{seg}$ corresponds to (\ref{oh-loss}),
   while the loss $L_{RNVP}$ corresponds to (\ref{rnvp-loss}).
   }
\label{semseg-setup}
\end{figure*}
\begin{myalign}\label{oh-loss}
  L_\mathrm{seg}(\theta) =  
    &- \sum_i \sum_j \llbracket \textbf{s}_{ij} = 0\rrbracket \, \mathrm{log}\, P_{\theta}(\textbf{y}_{ij }|\textbf{x}) 
    \nonumber\\ 
    + &
    \lambda \sum_i \sum_j \, \llbracket \textbf{s}_{ij} = 1\rrbracket \, 
      \mathrm{KL}(  U \, || \, P_{\theta}(\textbf{y}_{ij}|\textbf{x})) 
\;.
\end{myalign}
In the above equation,  $\textbf{y}$ represents the ground truth labels,
while $\textbf{s}$ represents the OOD mask 
where zeros correspond to unchanged pixels of the training image (inliers)
and ones denote the pasted RNVP output (outliers).

Figure \ref{semseg-setup} shows the proposed training setup
for dense open-set recognition.
We generate a synthetic outlier by sampling RNVP, 
and paste it at a random position in the training image.
Images with mixed content are given
to our discriminative model for dense prediction,
which optimizes the discriminative loss (\ref{oh-loss}).
Consequently, the gradients of (\ref{oh-loss})
are back-propagated to the RNVP.
Additionally, we maximize the likelihood of the image patch 
replaced by the generated outlier.
This is necessary in order to keep
the samples generated by RNVP at the border of the training distribution \cite{LeeLLS18}.
Due to the convenient architecture of RNVP
we are able to generate synthetic samples that vary in
spatial dimensions.
Hence, we can train our model with synthetic outliers of different sizes.
This is not straightforward to accomplish with other types of
generative models and we consider that as a distinct advantage of our approach.
The proposed procedure is summarized by Algorithm \ref{alg-optim}.

Semantic segmentation is known as a memory intensive task.
Hence, we optimize memory
consumption by using gradient checkpointing \cite{ChenXZG16,krevso2020efficient} 
which trades computation time for lower memory consumption.
We apply the checkpointing procedure 
both on our dense classifier and the RNVP.

\begin{algorithm}[b]
    \textbf{Require:} $\lambda > 0$\;
    \textbf{Define} RNVP: $\textbf{z} = \textbf{f}_{\theta_R}(\textbf{x}),  \textbf{x} = \textbf{f}_{\theta_R}^{-1}(\textbf{z})$\;
    \textbf{Define} Classifier: $P_{\theta_C}(\textbf{y}|\textbf{x})$\;
    \textbf{Define} Optimizers: $O_\mathrm{R}(\theta_R), O_\mathrm{C}(\theta_R)$\;
    \Repeat{convergence}{
      $\textbf{x}, \textbf{y} = \mathrm{obtain\_minibatch()}$\;
      $\textbf{z} = \mathrm{sample} \, N(0, I)$\;
      $\textbf{x}_{ood} = \textbf{f}_{\theta_R}^{-1}(\textbf{z})$\;
      $\textbf{x}, \textbf{x'}, \textbf{y}, \textbf{s} = \mathrm{process\_batch}(\textbf{x}, \textbf{x}_{ood}, \textbf{y})$\;
      $L_\mathrm{seg}(\theta_C) =  $\\$ \quad - \sum_i \sum_j \llbracket \textbf{s}_{i,j} = 0\rrbracket \, \mathrm{log}\,P_{\theta_C}(\textbf{y}_{i,j }|\textbf{x}) $\\$ \quad + \lambda \sum_i \sum_j \, \llbracket \textbf{s}_{i,j} = 1\rrbracket \,  \mathrm{KL}(  U \, || \, P_{\theta_C}(\textbf{y}_{i, j}|\textbf{x}))$\;
      $\theta_\mathrm{C} += O_\mathrm{C}.update(\nabla_{\theta_C} \, L_\mathrm{seg})$\;
      $L_\mathrm{RNVP} = -\mathrm{log}(p_{z}(\textbf{f}_{\theta_R}(\textbf{x}'))) $\\$ \quad - \mathrm{log}\left|det(\frac{\delta \textbf{f}_{\theta_R}(\textbf{x}')}{\delta \textbf{x}'})\right|$\;
      $\theta_\mathrm{R} += O_\mathrm{R}.update(\nabla_{\theta_R} \, L_\mathrm{RNVP} + \nabla_{\theta_C} \, L_\mathrm{seg})$\;
    }
 \caption{Dense open-set recognition classifier training.}
 \label{alg-optim}
\end{algorithm}

\subsection{Effects of Temperature Scaling onto OOD Detection}

The loss (\ref{oh-loss})
encourages high entropy of the softmax output in outlier pixels.
This improves the outlier detection performance
with respect to the standard cross-entropy loss.
However, OOD accuracy can be further improved by applying
temperature scaling during the inference phase \cite{GuoPSW17,LiangLS18}.
Dividing pre-softmax logits with a constant $T>1$
moves the softmax output of every sample 
closer (but not equally closer) to the uniform distribution.
We empirically show that such practice
yields more appropriate values of the scoring function $s$ 
and enables recognition of
some previously undetected outliers \cite{LiangLS18}.

\begin{table*}[b]
\caption{OOD detection performance of 
  the VGG13 model \cite{SimonyanZ14a} trained on the
  CIFAR10 dataset. The RNVP-based approach outperforms 
  both the max-softmax baseline \cite{HendrycksG17}
  and the GAN-based approach
  \cite{LeeLLS18} 
  across multiple OOD datasets and metrics. 
  Our RNVP-based approach achieves 85.98\% accuracy on
  the SVHN test set, while the GAN-based approach
  achieves 80.27\%.
  }
\label{iw-table}
\centering
\begin{tabular}{l|ccc}      
                                    OOD dataset           & TNR at TPR 95\% & AUROC  & OOD det. acc. \\ \hline
                                                                   & \multicolumn{3}{c}{Baseline / GAN outliers / RNVP outliers (ours)}                                \\ \hline
 SVHN                  & 14.0/12.7/\textbf{14.8}          & 46.2/46.2/\textbf{83.0}  & 66.9/65.9/\textbf{78.9}          \\
                                    LSUN (resize)         & 14.0/26.8/\textbf{46.5}           & 40.8/61.9/\textbf{87.5}  & 63.2/73.2/\textbf{79.3}          \\
                                    TinyImageNet (resize) & 14.0/28.1/\textbf{33.7}           & 39.8/66.2/\textbf{79.7}  & 62.9/73.2/\textbf{73.3}          \\ \hline
\end{tabular}
\end{table*}
We observe that dense classifiers tend to assign a low max-softmax
score in pixels at semantic borders.
Consequently, any of these pixels end up wrongly classified as outliers.
This happens because the border pixels typically have two dominant logits
(belonging to the two neighboring classes),
while the other logits have significantly smaller values.
On the other hand, undetected outlier pixels do not follow such pattern.
It is easy to show that temperature scaling influences more the max-softmax score
in pixels with homogeneous non-maximum logits than in pixels with two dominant logits.
This practice improves 
separation of OOD score between border and outlier pixels,
as well as the general OOD detection performance.

\section{\uppercase{A Novel dense OOD Detection Dataset}}

We propose a novel OOD detection dataset
which we obtained by relabeling the Mapillary Vistas \cite{NeuholdOBK17} dataset.
The original Vistas dataset consists of $18\,000$ training images 
and $2\,000$ validation images with 66 classes.
We propose to use human classes as outliers 
due to their dispersion across scenes 
and visual diversity from other objects.
We create a novel dense OOD dataset by excluding all images with class 
\textit{person} and the three rider classes to the test subset. 
Consequently, our dataset 
has $8\,003$ train images 
and $830$ validation images. 
The test set contains $11\,167$ images
($8\,003+830+11\,167 = 20\,000$).
We refer to our dataset as Vistas-NP (no persons)\footnote{ \url{https://github.com/matejgrcic/Vistas-NP}}.

The obtained dataset is similar to BDD-Anomaly \cite{bddanomaly} 
which selects the classes \textit{motorcycle} 
and \textit{train} classes as visual anomalies. 
However, the class \textit{motorcycle}
is often visually alike to class \textit{bike},
while the class \textit{train} is often 
visually similar to class \textit{bus}.
Therefore, the error of recognizing 
an OOD pixel on a motorcycle as a \textit{bike} 
receives an equal penalty 
as the error of recognizing that pixel as a \textit{person}.
We believe that choosing persons as outliers
is a more sensible choice
since the whole category is removed from the dataset.
Another advantage of Vistas dataset is better variety.
All images from BDD dataset originate from the USA. 
Contrary, the Mapillary Vistas dataset 
contains a more extensive set of world-wide driving scenes.

Table \ref{va-lf2} shows a comparison of the 
Vistas-NP test vs.\ FS Lost \& Found \cite{fs-lf} 
and BDD-Anomaly \cite{bddanomaly} test. 
FS Lost \& Found contains 100 publicly available images 
while BDD-Anomaly test set includes 361 images. 
Our test subset
has significantly more images 
with diverse instances of anomaly classes. 
Consequently, Vistas-NP is able to provide 
a more comprehensive insight into OOD detection performance.

\begin{figure*}[bp]
\centering
\begin{tabular}{@{}c@{\;}c@{\;}c@{\;}c@{}}
  \includegraphics[width=0.325\linewidth]{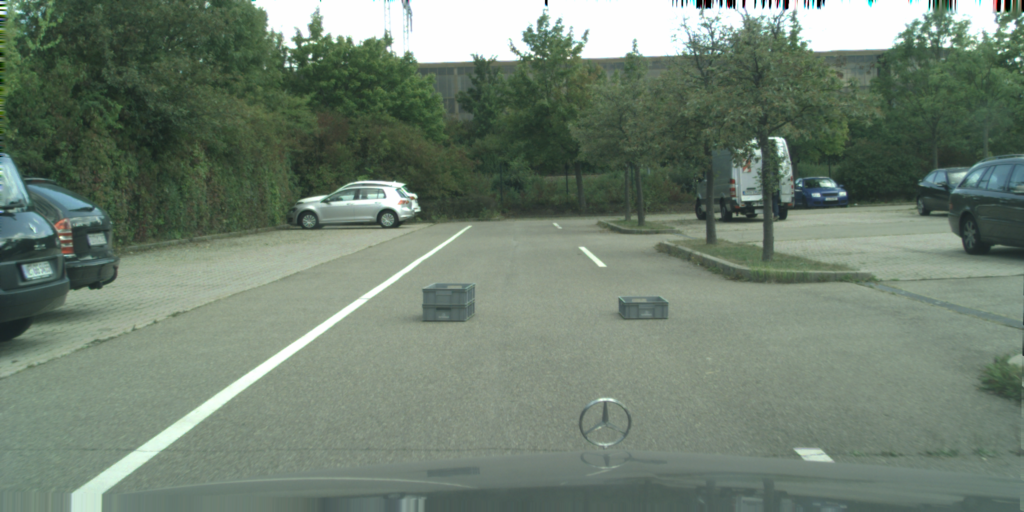} &
  \includegraphics[width=0.325\linewidth]{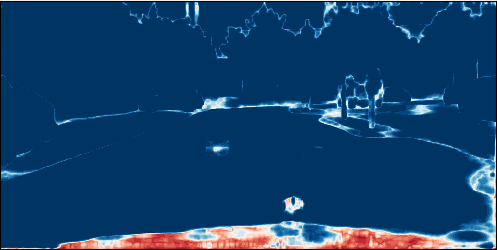} &
  \includegraphics[width=0.325\linewidth]{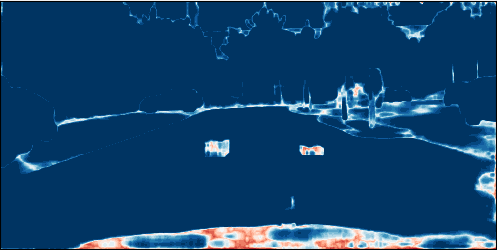} \\
   (a) & (b) & (c) \\
\end{tabular}
\caption{Model performance on FS Lost \& Found dataset \cite{fs-lf}. Figure (a) shows
the original image. Figure (b) shows the output of baseline model,
while figure (c) shows the output of the model trained in the proposed
setup. Our approach significantly improves the dense OOD detection performance.
}
\label{fig:lf-results}
\end{figure*}

\section{\uppercase{Experiments}}
We explore open-set recognition performance
of the proposed RNVP-based approach.
We first address the image-wide setup 
on CIFAR-10, 
where we compare the outlier detection performance
of our RNVP-based approach with 
the original GAN-based approach \cite{LeeLLS18} and the
max-softmax baseline \cite{HendrycksG17}.
Subsequently, we evaluate an adaptation of our approach
for dense prediction as proposed in \ref{sec:dense}.
We demonstrate effectiveness on
public open-set recognition datasets
(Lost \& Found, WD-Pascal, and StreetHazards)
as well as on the proposed novel dataset Vistas-NP.
\begin{table}[t]
\centering
\caption{
Comparison of Vistas-NP (ours) 
  with respect to FS Lost \& Found \cite{fs-lf}
  and BDD-Anomaly \cite{bddanomaly}. 
  Note that FS Lost \& Found recommends
  training on
  Cityscapes.}
 \label{va-lf2}
\begin{tabular}{cccc}
\hline
                             & Vistas-NP (ours)   & FS L\&F  & BDD-A \\ \hline
                             & \multicolumn{3}{c}{Label shares (\%)} \\ \hline
\multicolumn{1}{c|}{Inlier}  & 94.2              & 81.2     & 82.3   \\
\multicolumn{1}{c|}{Outlier} & 0.6               & 0.2      & 0.8    \\
\multicolumn{1}{c|}{Ignore}  & 5.2               & 18.6     & 16.9   \\ \hline
                             & \multicolumn{3}{c}{Number of images}       \\ \hline
\multicolumn{1}{c|}{Train}   & 8\,003              & 5\,000    & 6\,688   \\
\multicolumn{1}{c|}{Test}    & 11\,167              & 100      & 361    \\ \hline
\end{tabular}
\end{table}

\subsection{Image-wide OOD Detection}
We evaluate our image-wide open-set  recognition approach in experiments
with the VGG-13 backbone \cite{SimonyanZ14a} on CIFAR-10 \cite{cifar10kriz}.
We evaluate OOD detection performance 
using multiple threshold-free metrics \cite{HendrycksG17} 
on outliers from LSUN \cite{YuZSSX15} 
and Tiny-ImageNet.
We use the maximum softmax probability (MSP)
as a baseline.

We compare our RNVP-based method 
with the original formulation of this method, which is based on adversarial generative training \cite{LeeLLS18}.
We demonstrate advantage of RNVP-based setup
in experiments with the same setup
as proposed in \cite{LeeLLS18}.
Table \ref{iw-table} shows the resulting OOD detection performance.
Our RNVP-based approach 
outperforms other approaches across all metrics 
without losing the classification accuracy.
We train the classifier for 100 epochs with batch size 64.
In contrast to \cite{LeeLLS18}, we set the loss-modulation hyper-parameter
$\lambda = 1$ and do not validate 
it for a particular OOD dataset.
The employed RNVP model consists of 3 residual blocks with 32 feature maps
in every coupling layer.
Downsampling is performed twice.
RNVP's parameters are optimized with Adam optimizer.
The training lasts for approximately 10 hours on a single NVIDIA Titan Xp
GPU. Max memory allocation peaks at 1.3 GB with gradient checkpointing \cite{ChenXZG16} of RNVP and 2.3 GB without checkpointing.
Note that \cite{LeeLLS18} also reports results 
for a different setup
where outlier samples are sampled 
from external negative datasets.
Those results are not relevant 
in the scope of this work.

\subsection{Dense Open-set Recognition}

We consider a dense open-set recognition approach which
jointly trains a generative model of synthetic outliers as described in \ref{sec:dense}.
We use a Ladder-style DenseNet-121 \cite{krevso2020efficient} (LDN-121) on all datasets.
LDN-121 is chosen
due to its memory efficiency and prediction accuracy.
Note that the proposed procedure is 
independent from the particular dense prediction model.
We always train on random $512 \times 512$ crops which
we sample from images resized to 512 pixels (shorter edge).
We preserve the original label resolution for all datasets except Vistas-NP.
Labels of the Vistas-NP dataset 
are resized to 512 pixels. 
The output of LDN-121 is bilinearly upsampled 
to the matching label resolution.
During the training we use the batch size 6 
(this would not be possible without checkpointing)
and set $\lambda$ to $1*10^{-3}$.
The temperature scaling procedure uses $T=2$ for softmax entropy and
$T=10$ for max-softmax OOD score.
The value of $\lambda$ is chosen in a way that it does not affect
model's semantic segmentation performance on one held-out training image,
while the parameter $T$ is choosen so it gives the best OOD results on the held-out image.
Parameters of LDN-121 are optimized using Adam optimizer with learning rate $1*10^{-4}$
for backbone parameters and $4*10^{-4}$ for upsampling path.
For LDN-121 we decay learning rate throughout epochs
using cosine annealing procedure to minimal value of $1*10^{-7}$.
Additionally, LDN's backbone is initialized with ImageNet weights.
Architecture of RNVP consists of 3 residual blocks with 32 feature maps
in every coupling layer.
Downsampling is performed three times.
Parameters of RNVP are optimized using Adam with default hyperparameters.
The generated outliers have spatial dimensions uniformly
selected from the set $\{64, 72, 80, 88, 96, 104, 112, 120, 128, 136, 144\}$.
Consequently, each outlier takes 1.5-8\% of the image area.

We demonstrate general applicability of the proposed method
by training it on Cityscapes \cite{CordtsORREBFRS16}
and testing dense outlier detection
performance on Fishyscapes Lost and Found \cite{fs-lf}.
We train all models for 54k iterations.
We test the contribution of our jointly trained model by substituting synthetic outliers with Gaussian noise. We refer to this baseline as LDN + noise.
Table \ref{lf-perf} shows significant improvement
of the proposed approach with respect to both baselines.
The first two columns show the tested model and the achieved mIoU accuracy
on the Cityscapes validation subset.
The last two columns show OOD detection performance, where
AP stands for average precision while
F95 stands for TPR at FPR 95\%.
We show the performance of our models when outliers are
detected using the max-softmax probability (MSP) and the entropy of softmax output (H).

\begin{figure*}[bp]
\centering
\begin{tabular}{@{}c@{\;}c@{\;}c@{\;}c@{}}
  \includegraphics[width=0.325\linewidth]{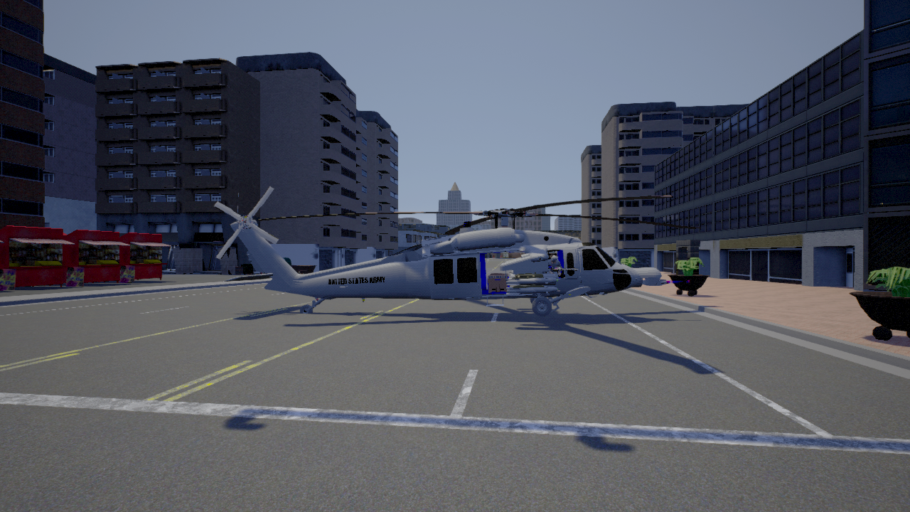} &
  \includegraphics[width=0.325\linewidth]{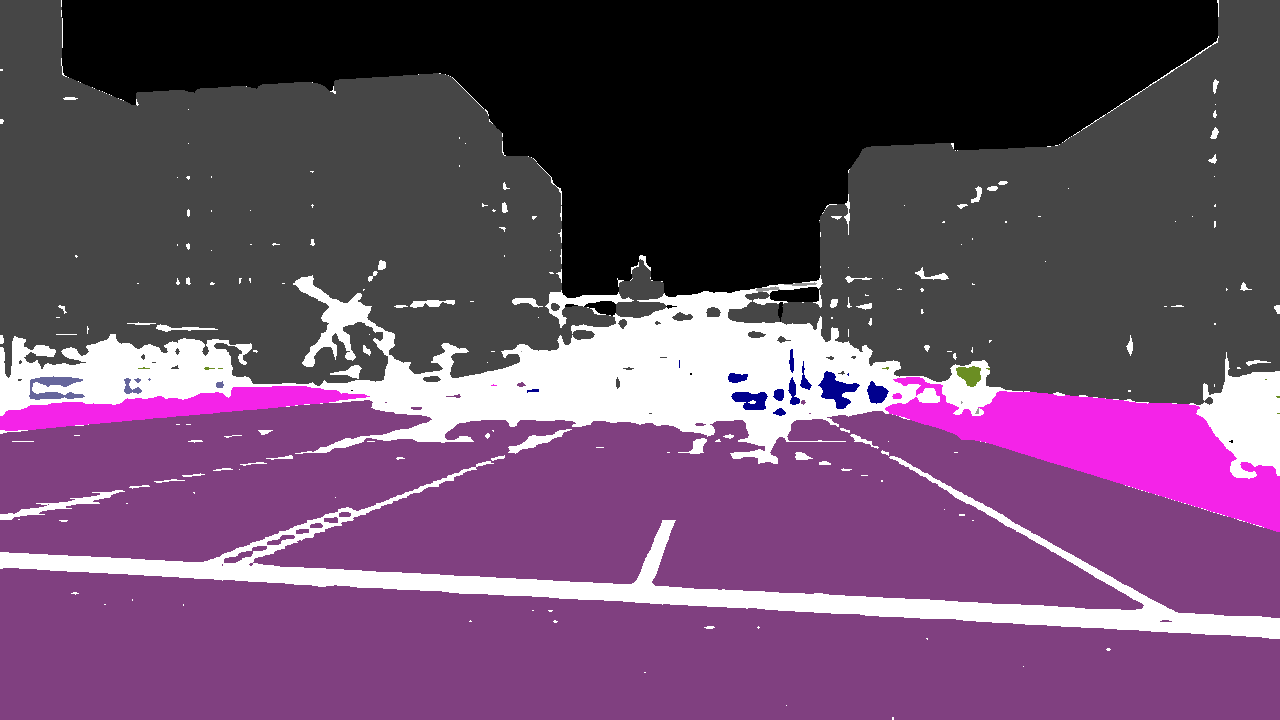} &
  \includegraphics[width=0.325\linewidth]{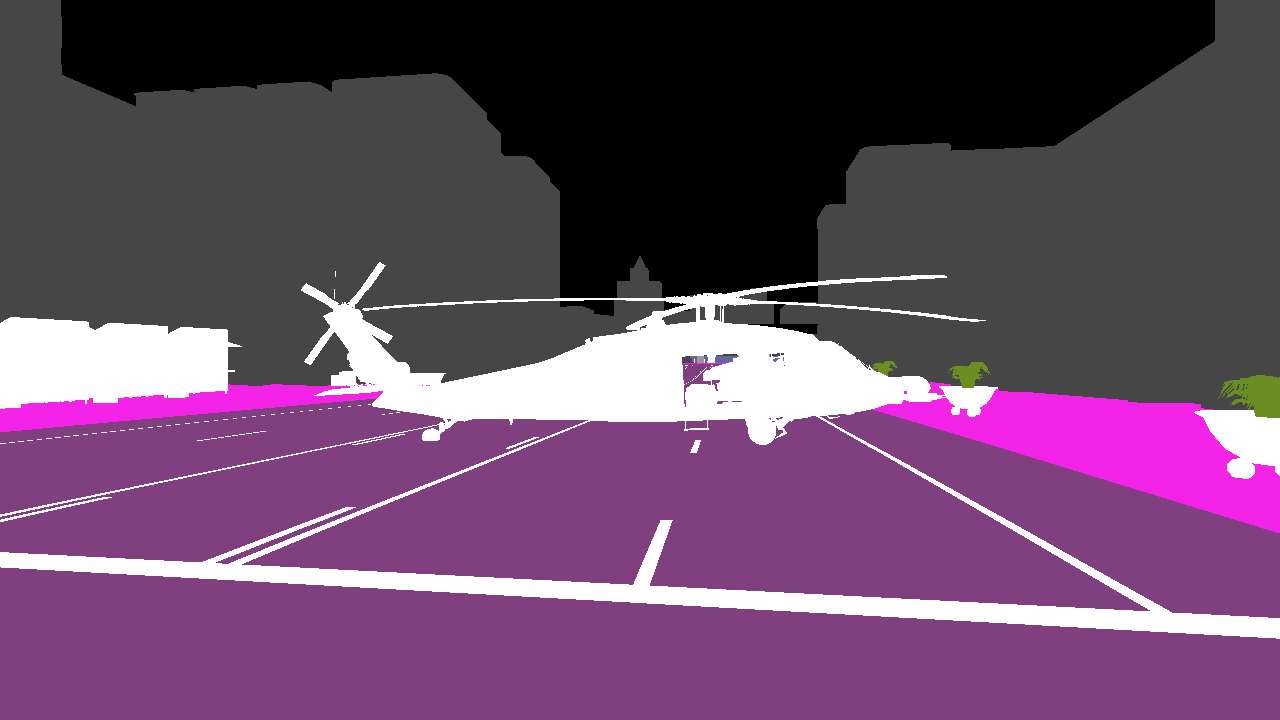} \\
   (a) & (b) & (c) \\
\end{tabular}
\caption{
Qualitative results on a StreetHazard test image
in which the outlier object corresponds to a helicopter.
We show the input image (a),
dense open-set prediction by the proposed method (b),
and the ground-truth labels (c).
Outlier objects are marked in white.
A pixel is marked as outlier if the corresponding
max-softmax OOD score is higher than 80\%.
}
\label{fig:caos-res}
\end{figure*}

\begin{table}[h]
\centering
\caption{
  Dense open-set recognition on Fishyscapes Lost \& Found \cite{fs-lf} with 
  LDN-121 \cite{krevso2020efficient}.
  Models are trained on Cityscapes dataset \cite{CordtsORREBFRS16}.
  SO denotes synthetic outliers.
  AP stands for average precision, while F95 represents
  TPR at FPR 95\%.}
  \label{lf-perf}
\begin{tabular}{l|c|cc}
\multicolumn{1}{l|}{} & Cityscapes  & \multicolumn{2}{c}{FS L\&F}   \\ \hline
Model                 & mIoU          & AP $\uparrow$          & F95 $\downarrow$     \\ \hline
LDN, MSP (baseline) & \textbf{72.1}          &     3.9      & 30.8 \\
LDN + noise & 71.3 & 4.9 & 27.0 \\ \hline
LDN+SO, T=1, MSP        & 71.5 & 6.8 & 26.8         \\ \hline
LDN+SO, T=1, H        & 71.5 & 12.5 & 26.0        \\
LDN+SO, T=10, MSP        & 71.5 & \textbf{16.5} & \textbf{23.3}         \\
\end{tabular}
\end{table}

Figure \ref{fig:lf-results} shows qualitative results on FS Lost \& Found.
Ideally, OOD pixels should
be painted in red which signals low-confidence predictions.
The baseline fails to detect two boxes at the road as anomalies, while
the proposed model performs much better.

Our training lasts for 94k iterations.
We also test the LDN-121 using the proposed Vistas-NP dataset.
As before, our OOD detection baseline is a discriminatively trained closed-set
model activated with max-softmax.
Table \ref{semseg-vnp}
shows the resulting dense open-set recognition performance.
The first two columns correspond to the 
mIoU and the AP performance on the test split.
The last column corespons to AP performance on
WD-Pascal\footnote{\url{https://github.com/pb-brainiac/semseg_od}} \cite{bevandic19simul}.
Note that WD-Pascal contains people marked as inliers.
However, there are only few images with people so they do not affect
average precision score significantly.
The bottom section of the table 
illustrates advantages of 
softmax entropy and temperature scaling.

\begin{table}[h]
\centering
\caption{Dense open-set recognition with 
  LDN-121 \cite{krevso2020efficient} trained on the Vistas-NP dataset.
  Our model improves dense OOD detection
  on both Vistas-NP test set and WD-Pascal \cite{bevandic19simul}
  without impairing the segmentation accuracy.
  }
  \label{semseg-vnp}
\begin{tabular}{l|cc|c}
                   & \multicolumn{2}{c|}{Vistas-NP} & WD-Psc.  \\ \hline
Model              & mIoU       & AP            & AP           \\ \hline
LDN, MSP (baseline) & 61.5 &    8.6    &   7.0       \\ \hline
LDN+SO, T=1, MSP     &  61.6   & 9.3 & 14.1\\ \hline
LDN+SO, T=1, H     &      61.6      & 13.7 & 17.8 \\
LDN+SO, T=10, MSP     &      61.6      & 16.2 & 20.5\\
LDN+SO, T=2, H     &      \textbf{61.6}      & \textbf{16.9} & \textbf{21.5}\\
\end{tabular}
\end{table}

\begin{table*}[h]
\centering
\caption{
Results on StreetHazards \cite{bddanomaly} dataset.
Our method equals the best AP score, achieves the best AUROC score
and the third best FPR95.
Additionally, we achieve the best mIoU accuracy.
}
\label{sh-res}
\begin{tabular}{l|cccc}
Model & mIoU $\uparrow$ & AP  $\uparrow$          & FPR95   $\downarrow $     & AUROC  $\uparrow $      \\ \hline
LDN, MSP \cite{HendrycksG17}  & 56.2     &  7.3           & 30.8          & 89.0          \\
Dropout \cite{GalG16}\cite{Yingda2020} & /  &  7.5           & 79.4          & 69.9          \\
AE \cite{BaurWAN18}\cite{Yingda2020} &  /    &  2.2           & 91.7          & 66.1          \\
MSP + CRF \cite{bddanomaly} & /    &  6.5       & 29.9          & 88.1          \\ 
SynthCP, t=1 \cite{Yingda2020} & /   &  8.1           & 46.0          & 81.9          \\
SynthCP, t=0.999 \cite{Yingda2020} & /   & 9.3           & 28.4          & 88.5          \\ 
Ensemble OVA \cite{Franchi2020}&  54.0 & \textbf{12.7}  & \textbf{21.9} & 91.6\\
OVNNI \cite{Franchi2020}& 54.6  & 12.6  & 22.2 & 91.2 \\ \hline
LDN + SO, T=1, MSP  & \textbf{59.7}   &  8.6           & 26.1 & 90.2          \\
LDN + SO, T=10, MSP & \textbf{59.7} & 12.1 & 29.1          & 90.8 \\
LDN + SO, T=1, H & \textbf{59.7} & 11.3 & 25.7         & 91.1 \\
LDN + SO, T=2, H & \textbf{59.7} & \textbf{12.7} & 25.2         & \textbf{91.7} \\
\end{tabular}
\end{table*}

Finally, we evaluate the proposed method on the StreetHazards \cite{bddanomaly} dataset.
The dataset contains 12 training classes. As proposed in \cite{bddanomaly}, we calculate
average precision for every image and report the mean value. The model is trained for
43k iterations.
Table \ref{sh-res} presents the obtained results
and compares it with the previous work.
We achieve the best AUROC score, equal the best AP score, and
outperform all previous approaches
with respect to segmentation accuracy by a wide margin.
Note that the best method \cite{Franchi2020} uses ensemble learning.

Figure \ref{fig:caos-res} shows results of LDN-121 trained in the proposed
procedure. We mark pixels as outliers if the max-softmax score is higher than 80\%.
Parameter $T$ is set to 10.

The proposed procedure consumes different amounts of GPU memory depending on 
the spatial dimensions of generated outliers.
We asses the memory consumption using NVIDIA Titan Xp and batch size 4.
We measure memory allocation of 9.61 GB for maximal outlier size,
while the minimal outlier size consumes 7.88 GB of memory.
When we apply the gradient checkpointing, memory allocation peaks at 5.55 GB,
while the minimal allocation equals to 4.92 GB.
Information about the GPU memory allocation is obtained using $torch.cuda.max\_memory\_allocated()$.

\section{\uppercase{Conclusion}}

We have presented a novel dense open-set recognition approach
based on discriminative training with jointly trained synthetic outliers.
The synthetic outliers are obtained by sampling 
a generative model based on normalized flow
that is trained alongside a dense discriminative model
in order to produce samples at the border 
of the training distribution.
We paste the generated samples into 
densely annotated training images,
and learn dense open-set recognition models
which perform simultaneous semantic segmentation
and dense outlier detection.
Experiments on CIFAR-10 show 
that synthetic outliers generated by RNVP 
lead to better open-set performance
then their GAN counterparts.
We present dense open-set recognition experiments 
on a novel dataset which we call Vistas-NP,
as well as on three public datasets
which were proposed in the prior work.
The proposed approach is competitive with
respect to the state of the art on the StreetHazards dataset.
Additionally, we outperform baselines on two other
dense OOD detection datasets.
Suitable avenues for future work include
increasing the capacity of the generative model,
combining the proposed approach with 
noisy outliers from some large general-purpose dataset,
and devising more involved approaches
for simultaneous discriminative and generative modeling.

\section*{\uppercase{Acknowledgments}}
\noindent
We thank Ivan Grubišić, Marin Oršić and Jakob Verbeek on their useful comments.
This work has been supported by the European Regional Development Fund under the project "A-UNIT - Research and development of an advanced unit for autonomous control of mobile vehicles in logistics" (KK.01.2.1.02.0119).
\bigskip
\bigskip




\bibliographystyle{apalike}
{\small
\bibliography{example}}

\end{document}